\documentclass[11pt]{article}
\usepackage{hyperref}
\usepackage{txfonts}

\usepackage{graphicx}
\graphicspath{{./Fig/}}
\usepackage{authblk}

\title{Exhaustive search for sparse variable selection \\in linear regression}

\author{Yasuhiko~Igarashi$^1$, 
Hikaru~Takenaka$^2$, 
Yoshinori~Nakanishi-Ohno$^3$,
Makoto~Uemura$^4$,
Shiro~Ikeda$^5$, and 
Masato~Okada$^{1,2}$\thanks{okada@k.u-tokyo.ac.jp}}

\affil[1]{Research and Service Division of Materials Data and Integrated System, 
National Institute for Material Science, 1-2-1 Sengen, Tsukuba, Ibaraki, 305-0047, Japan}
\affil[2]{Graduate School of Frontier Sciences, The University of Tokyo, 5-1-5, Kashiwanoha, Kashiwa, Chiba, 277-8561, Japan }
\affil[3]{Graduate School of Arts and Sciences, The University of Tokyo, 3-8-1, Komaba, Meguro-ku, Tokyo 153-8902, Japan}
\affil[4]{Hiroshima Astrophysical Science Center, Hiroshima University, 1-3-1, Kagamiyama, Higashi-Hiroshima, Hiroshima 739-8526, Japan}
\affil[5]{The Institute of Statistical Mathematics, 10-3, Midori-cho, Tachikawa, Tokyo 190-8562, Japan}

\begin{document}
\maketitle
\begin{abstract}
We propose a $K$-sparse exhaustive search (ES-$K$) method and a $K$-sparse approximate exhaustive search method (AES-$K$) 
for selecting variables in linear regression. 
With these methods, 
$K$-sparse combinations of variables are tested exhaustively 
assuming that the optimal combination of explanatory variables is $K$-sparse. 
By collecting the results of exhaustively computing ES-$K$, 
various approximate methods for selecting sparse variables can be summarized 
as density of states. 
With this density of states, we can compare different methods for selecting sparse variables such as relaxation and sampling. 
For large problems where the combinatorial explosion of explanatory variables is crucial, 
the AES-$K$ method enables density of states to be effectively reconstructed 
by using the replica-exchange Monte Carlo method and the multiple histogram method. 
Applying the ES-$K$ and AES-$K$ methods to type Ia supernova data, we confirmed the conventional understanding in astronomy 
when an appropriate $K$ is given beforehand. 
However, we found the difficulty to determine $K$ from the data. 
Using virtual measurement and analysis, we argue that this is caused by data shortage.
\end{abstract}
%

\section{Introduction}
\label{sec:intro}
Selecting the variables of a linear regression model is a fundamental problem of statistics. 
When there are $N$ explanatory variables, 
 the simplest method for selecting variables is to exhaustively search for all combinations, which requires 
the combinations of variables to be estimated 
$2^N-1 = {}_N\mathrm{C}_1 + {}_N\mathrm{C}_2 + \dots +{}_N\mathrm{C}_K \dots + {}_N\mathrm{C}_N$ times 
the number of estimations that the combinations of variables requires times 
\cite{Ichikawa2014,Kuwatani2014,Nagata2015,Igarashi2016}. 
We call this naive method the ``exhaustive search (ES) method'' \cite{Igarashi2016}. 
Cover and Van Campenhout reported that any exact methods for variable selection 
come at the expense of a computational complexity of at least $O(2^N)$ \cite{Cover1977}, and this is true for the ES method as well. 

It is easy to imagine that the ES method becomes intractable for a large $N$. To reduce the computational load, sampling methods are effective. A sampling method for variable selection using the Markov chain Monte Carlo (MCMC) method \cite{Metropolis1953} was proposed in 1993 \cite{Picard1984,Shao1993,George1993}. Later, using the replica exchange Monte Carlo (REMC) method \cite{Hukushima1996}, also known as parallel tempering \cite{Geyer1991}, a more efficient sampling method for variable selection was proposed \cite{Kim2006}. Using the multiple histogram method \cite{Ferrenberg1989} in combination with the REMC method, Nagata \textit{et al.} proposed a method for estimating the density of states and applied it to variable selection \cite{Nagata2015}. We call this method the ``approximate exhaustive search (AES) method'' \cite{Igarashi2016}. 

In this paper, we consider sparse variable selection for linear regression. This is an important method, especially when the number of explanatory variables is larger than that of the data dimensions but the number of true variables is small. Generally, there are two approaches to sparse variable selection. One is the relaxation approach, such as the least absolute shrinkage and selection operator (LASSO) method using an $L1$-norm regularization term \cite{Tibshirani1996,Candes2005,Donoho2006b,Donoho2006,Candes2006}. The other is the sampling approach, which uses MCMC or REMC \cite{Shao1993,George1993,Kim2006}. We should emphasize that the sampling approach has been used mainly in order to derive the optimal solution, but it can also be used to estimate the density of explanatory variable combinations with respect to any performance measure. 

In this study, we extend the ES and AES methods to sparse variable selection in linear regression. Assuming that the optimal combination of explanatory variables is $K$-sparse, i.e., it has $K$ non-zero explanatory variables, we propose a $K$-sparse exhaustive search (ES-$K$) method in which $K$-sparse combinations are exhaustively searched. 
In the case of a large $K$, we propose a $K$-sparse approximate exhaustive search (AES-$K$) method. 
The typical settings of the model selection problem are to find an optimal set of variables. 
However, there might be multiple models that behave similarly or have a smaller generalization error. We tested the reliability of the conventional model with the ES-$K$ method. We confirmed that the models selected on the basis of free energy (FE) and cross validation error (CVE) include these two variables. We analyzed the data of type Ia supernovae in the Berkeley Supernova Database \cite{Silverman2012} with the ES-$K$ and AES-$K$ methods. The widely accepted model is that the absolute magnitude at maximum depends on the color and light-curve width \cite{Uemura2015}. 

The rest of this paper is organized as follows. In Section 2, we formulate sparse variable selection in the linear regression problem and explain the ES-$K$ and AES-$K$ methods. In Section 3, we analyze type Ia supernova data using the the methods. In Section 4, we conduct virtual measurement and analysis to discuss the results. In Section 5, we conclude this paper. 

\section{Methods}
\subsection{Exhaustive search method}
\label{subsec:ES}
Here, we describe the exhaustive search (ES) method for the linear regression problem. Let us suppose that an objective variable $y_\mu$ is well explained with some of the $N$ explanatory variables $\mathbf{x}_\mu = (x_{\mu 1},x_{\mu 2},\dots,x_{\mu N})^\mathrm{T}$ through a linear measurement process. Given a dataset comprising $p$ samples of $\mathbf{y} = (y_1,y_2,\dots,y_p)^\mathrm{T}$ and $\mathbf{X}=(\mathbf{x}_1,\mathbf{x}_2,\dots,\mathbf{x}_p)^\mathrm{T}$, we can write $\mathbf{y} = \mathbf{1}\beta_0 + \mathbf{X} \mathbf{\beta} + \mathbf{\epsilon}$, where $\mathbf{1}$ is a $p$-dimensional all-one vector, $\beta_0$ is a constant coefficient, $\mathbf{\beta} = (\beta_1,\beta_2, \dots, \beta_N)^\mathrm{T}$ is a coefficient vector of $\mathbf{X}$, and $\mathbf{\epsilon}$ is a measurement noise vector. We assume that the measurement noise is Gaussian, where the variance is $\sigma^2_\mu$ ($\mu=1,\dots,p$) and known. 
The goal is to estimate the coefficients of linear regression. Although the maximum likelihood estimate of $\beta_0$ and $\mathbf{\beta}$ is obtained by the weighted least squares for $p\ge N$, it cannot be applied for $p < N$, which is the case we consider in this article. Assuming that $\mathbf{\beta}$ is sparse, namely, $\mathbf{\beta}$ has a small number of non-zero elements, we estimate the $\beta_0$ and $\mathbf{\beta}$ of the indeterminate linear equation. With the ES method, whether each coefficient $\beta_i$ is zero or not is determined by exhaustively evaluating all combinations of $N$ explanatory variables in terms of a certain information criterion, and subsequently, each value of non-zero $\beta_i$ is determined by the least squares method. The total number of all the combinations to be searched is $2^N-1$, and this is why the ES method requires a computational complexity of $O(2^N)$ \cite{Cover1977}. 

We formulate the linear regression problem by using an indicator variable that represents a combination of non-zero explanatory variables. The indicator is defined as an $N$-dimensional binary vector,
\begin{eqnarray}
\mathbf{c} =(c_{1}, c_2, \dots, c_{N}) \in \{0,1\}^N. 
\end{eqnarray}
Each variable $c_{i}$ takes $0$ or $1$: $c_{i}=1$ if the $i$th variable belongs to the combination and $c_{i}=0$ if it does not. Using the indicator $\mathbf{c}$, we can write the linear regression problem as 
\begin{eqnarray}
\mathbf{y} =  \mathbf{1}\beta_0 + \mathbf{X} (\mathbf{c} \circ \mathbf{\beta}) + \mathbf{\epsilon}, 
\label{eq:ESLR1}
\end{eqnarray}
where the symbol $\circ$ represents the Hadamard product, namely, $(\mathbf{c}\circ\mathbf{\beta})_i=c_i\beta_i$. 
This formulation makes the essence of the problem more explicit, and the best $\mathbf{c}$ for modeling and predicting 
an objective variables $\mathbf{y}$ is searched by minimizing the FE and the CVE with the ES method.  

\subsubsection{Free energy}
\label{subsub:FE}
The FE is an information criterion for selecting models in the framework of Bayesian inference. 
It is often replaced by an asymptotic formula called the Bayesian information criterion (BIC) \cite{Schwarz1978} due to its intractability in exact calculation. 
However, in our case, it can be analytically calculated, and its minimization is possible in sparse variable selection. 

Here, we derive the FE in linear regression. To select a combination of explanatory variables, $\mathbf{c}$, we calculate a posterior probability $P(\mathbf{c}|\mathbf{y})$, and the combination of explanatory variables that has the highest posterior probability can be regarded as the optimal model. According to Bayes' theorem, the posterior probability is given by 
\begin{eqnarray}
P(\mathbf{c}|\mathbf{y}) 
=\frac{P(\mathbf{y}|\mathbf{c})P(\mathbf{c}) }{P(\mathbf{y})} 
\propto P(\mathbf{y}|\mathbf{c}).
\label{eq:modelpost}
\end{eqnarray}
where the uniform prior probability $P(\mathbf{c})$ is used. 
In this case, the posterior probability is proportional to a marginalized likelihood function defined by
\begin{eqnarray}
P(\mathbf{y}| \mathbf{c})
= \int P(\mathbf{y}| \mathbf{\beta},\mathbf{c})
P(\mathbf{\beta}|\mathbf{c}) d\mathbf{\beta}.
\label{eq:P_marginalization}
\end{eqnarray}
The negative logarithm of the marginalized likelihood function is called the ``FE,'' namely, 
$\mathrm{FE} (\mathbf{c}) \equiv -\log P(\mathbf{y}|\mathbf{c})$, 
and the FE minimization is identical to the posterior probability maximization. 
Here, we assume that 
in the case of $c_i=1$, $P(\beta_i|c_i=1)$ is Gaussian distribution 
where the mean and variance are $0$ and $s$, respectively as follows. 
\begin{eqnarray}
P(\beta_i|c_i=1)=\frac{1}{\sqrt{2\pi s^2}}\exp\left(-\frac{\beta_i^2}{2s^2}\right),  \;
P(\beta_i | c_i=0) = \delta(\beta_i) \; 
(i=1,2,\dots,N). 
\end{eqnarray}
We estimate the variance $s$ using the observed data as described in the next paragraph. 
The likelihood function, $P(\mathbf{y}| \mathbf{\beta},\mathbf{c})$, is given by
\begin{eqnarray}
P(\mathbf{y}| \mathbf{\beta},\mathbf{c}) 
=
 \frac{1}{\mathrm{det}(2\pi\Sigma)^{1/2}} 
\exp
\left(-\frac{1}{2}\Delta^{\mathrm{T}}\Sigma^{-1}\Delta\right), 
\end{eqnarray}
where 
$\Delta = \left[ \mathbf{y}  - \left\{\mathbf{1}\beta_0 + \mathbf{X} (\mathbf{c} \circ \mathbf{\beta})
\right\}\right]$ 
and 
$\Sigma$ represents the covariance matrix of measurement noise, 
whose elements are given by $\Sigma_{ii}=\sigma^2_i$ ($i=1,\dots,p$) and $\Sigma_{ij} = 0$ ($i\neq j$). 
After a straightforward calculation, the resultant formula of the FE up to a constant is given as 
\begin{eqnarray}
\mathrm{FE}(\mathbf{c})
=\frac{p}{2}\log(2\pi) + \frac{1}{2}\log\mathrm{det}(\Sigma^2) +K \log s 
+ \frac{1}{2}\mathbf{y}^\mathrm{T}\Sigma^{-2}\mathbf{y} 						                       
 -\frac{1}{2} \mathbf{\mu}^\mathrm{T}\Lambda^{-1}\mathbf{\mu}
						- \frac{1}{2} \mathrm{det} (\Lambda)
\label{FE_c}
\end{eqnarray}
where we set
$\Lambda = (\mathbf{X}_\mathrm{I}^\mathrm{T}\Sigma^{-2}\mathbf{X}_\mathrm{I} +\frac{1}{s^2}\mathrm{I})^{-1}$ 
and $\mathbf{\mu}=\Lambda \mathbf{X}_\mathrm{I}^\mathrm{T}\Sigma^{-2}\mathbf{y}$, and 
$\mathbf{X}_{\mathrm{I}}$ is a matrix composed of non-zero explanatory variables. 
With the ES method, the FE is calculated for all combinations of explanatory 
variables and the combination minimizing the FE is taken as the optimal one. 

To estimate the optimal prior parameter,$s$, which maximizes the free energy $\mathrm{FE}(\mathbf{c},s)$
we derive the partial derivative of $\mathrm{FE}(\mathbf{c},s)$ with respect to $s$. 
When we set $z=\frac{1}{s^2}$, the partial derivative of $\mathrm{FE}(\mathbf{c},z)$ can be calculated as 
\begin{eqnarray}
\frac{\partial \mathrm{FE}(\mathbf{c},z)}{\partial z} 
=
-\frac{K}{2z} + \frac{1}{2}\mathbf{\mu}^\mathrm{T}\mathbf{\mu} 
+\frac{1}{2}\mathrm{Tr}(\Lambda). 
\end{eqnarray}
For finding relative minima of $\mathrm{FE}(\mathbf{c},z)$, 
we set $\frac{\partial \mathrm{FE}(\mathbf{c},z)}{\partial z} =0$ and derive the following self-consistent equation. 
\begin{eqnarray}
z = K\left(
\mathbf{\mu}^\mathrm{T}\mathbf{\mu} + \sum_{k=1}^K\frac{1}{b_k + z}
   \right)^{-1}
   \label{eq:z_self}
\end{eqnarray}
where $b_k$ represents the $k$-th eigen value of $\mathbf{\mu}^\mathrm{T}\Lambda^{-1}\mathbf{\mu}$. 
After many iterations of solving Eq. (\ref{eq:z_self}), we estimate the optimal prior parameter, $s(=\frac{1}{\sqrt{z}})$. 
When you set a sufficiently large variance $s$ and take no account of the third therm of Eq. \ref{FE_c}, 
you can calculate the FE for the uniform prior of $\mathbf{\beta}$. 

\subsubsection{Cross validation error}
\label{subsubsec:CV}
The performance of an explanatory variable combination can also be evaluated by using the CVE from the viewpoint of prediction error. The CVE asymptotically approaches the Akaike's Information criterion (AIC) \cite{Stone1977}. Specifically, we explain the $M$-fold CV used in this study. First, we randomly divide the indexes of data from $\mu=1$ to $p$ into $M$ parts, $B_1,\dots,B_M$. Next, for each $m$ ($=1,\dots,M$), we estimate the coefficients to obtain $\hat{\mathbf{\beta}}_m$ by using training data $y_\mu$ and 
$\mathbf{x}_{\mu}$ of $\mu\not\in B_m$. Finally, we calculate the CVE that measures the distance from the validation data $y_\mu$ and $\mathbf{x}_{\mu}$ of $\mu\in B_m$ to the trained coefficients $\hat{\mathbf{\beta}}_m$. The CVE is defined as a weighted mean squared error, 
\begin{eqnarray}
\mathrm{CVE}(\mathbf{c}) &=& \frac{1}{M} \sum_{m=1}^M \mathrm{CVE}_m(\mathbf{c}), \label{eq:CVE1}\\
\mathrm{CVE}_m(\mathbf{c}) &=& \frac{\sum_{\mu\in B_m}(y_{\mu}-\hat{y}_{\mu})^2/\sigma_{\mu}^2}{\sum_{\mu\in B_m}1/\sigma_{\mu}^2},
\label{eq:CVE2}
\\
\hat{y}_{\mu} &=& \mathbf{x}_{\mu}^\mathrm{T}(\mathbf{c}\circ\hat{\mathbf{\beta}}_{m}).
\label{eq:CVE3}
\end{eqnarray}

\subsection{$K$-sparse exhaustive search method}
The computational complexity of the ES method rises exponentially with the amount of data.
To overcome this problem, we propose a $K$-sparse exhaustive search (ES-$K$) method. 
The ES-$K$ method is based on the assumption 
that the optimal combination of explanatory variables is $K$-sparse, namely, 
$K$ components of $\mathbf{c}$ are explanatory variables. 
The ES-$K$ method searches the optimal $K$-sparse combination
for sparse variable selection in linear regression by calculating the FE and CVE of all $K$-sparse combinations. 

\subsection{$K$-sparse approximate exhaustive search method}
Even with the ES-$K$ method, the computational cost of $O(_N\mathrm{C}_K)$ is still relatively large. 
In this study, we focused on the REMC method \cite{Hukushima1996}, 
which is also used in the AES method \cite{Nagata2015,Igarashi2016}. 
The REMC method enables us not only to effectively derive a combination of explanatory variables 
with a minimal FE or CVE \cite{George1993,Kim2006} 
but also to estimate the density of states corresponding to the FE or CVE by combining it with the multiple histogram method \cite{Ferrenberg1989}. 
We called this the ``$K$-sparse approximate exhaustive search (AES-$K$) method.'' 

\subsubsection{Replica exchange Monte Carlo method}
The purpose of the REMC method is to efficiently sample $\mathbf{c}$ 
from the following Boltzmann distribution with energy $E$, which represents the FE or CVE. 
\begin{eqnarray}
P_\omega(\mathbf{c}|T_{\omega})=
\frac{1}{Z_\omega}
\exp\left(-\frac{E(\mathbf{c})}{T_{\omega}}\right), 
\end{eqnarray}
where $T_\omega>0$ is a ``temperature" parameter, and $Z_\omega$ is a partition function. For the REMC method, we prepare replicas of the Boltzmann distribution $P_\omega$ with several temperatures $0<T_1<\dots<T_\omega<\dots<T_\Omega$. The REMC method is used to sample a combination of explanatory variables $\mathbf{c}$ from the following joint probability distribution. 
\begin{eqnarray}
P(\mathbf{c}_1,...,\mathbf{c}_\Omega)=\prod_{\omega=1}^{\Omega}P_\omega(\mathbf{c}_\omega |T_\omega). 
\end{eqnarray}
The REMC method uses two sequential state transitions for sampling. 
\begin{enumerate}
\item
For each temperature, $\mathbf{c}$ is sampled in parallel from $P_{\omega}(\mathbf{c} |T_\omega)$ by using the Metropolis algorithm \cite{Metropolis1953}, in which the number of explanatory variables is fixed at $K$ \cite{Nakanishi2016}. 
\item 
The samples are exchanged between neighboring replicas, that is, between $\mathbf{c}_\omega$ and $\mathbf{c}_{\omega+1}$, with a probability of $\min\{1,r'\}$, where 
\begin{eqnarray}
r'&=&\frac{P_{\omega}(\mathbf{c}_{\omega+1}|T_\omega)P_{\omega+1}(\mathbf{c}_\omega |T_{\omega+1})}{P_{\omega}(\mathbf{c}_{\omega}|T_\omega)
P_{\omega+1}(\mathbf{c}_{\omega+1}|T_{\omega+1})}\nonumber\\
&=&\exp\left\{(1/T_{\omega+1}-1/T_{\omega})[E(\mathbf{c}_{\omega+1})-E(\mathbf{c}_{\omega})]\right\}. 
\end{eqnarray}
\end{enumerate}
After many iterations of these two steps, the obtained distributions of $\mathbf{c}$ converge to the joint probability distribution $\prod_{\omega=1}^{\Omega}P_{\omega}(\mathbf{c}_\omega |T_\omega)$. Exchanging the samples between different temperatures not only promotes convergence but also enables a search for the global optimal solution even though the energy has many local minima. 
 
\subsubsection{Multiple histogram method}
By combining the multiple histogram method \cite{Ferrenberg1989} with the REMC method, we can approximately estimate the density of states corresponding to the FE or CVE. Given histograms $H_\omega(E)$ of the $E$ obtained with the REMC method for various temperatures $T_\omega$, we can express the density of states $g(E)$ as 
\begin{eqnarray}
g(E)&=&\frac{\sum_{\omega=1}^{\Omega}H_\omega(E)}{\sum_{\omega=1}^{\Omega}n_{\omega}
\exp{(f_\omega-E/T_\omega)}}, 
\label{gE}
\end{eqnarray}
where $n_\omega$ is the total number of samples obtained at $T_\omega$, and $f_\omega$ is called the ``free energy,'' not to be confused with the FE defined in \S \ref{subsub:FE} as an information criterion, defined by 
\begin{eqnarray}
f_\omega&=&-\log\sum_{E}g(E)\exp{(-E/T_\omega)}.
\label{expf} 
\end{eqnarray}
Equations (\ref{gE}) and (\ref{expf}) are alternately solved by substituting them into each other to estimate the density of states $g(E)$. 

\subsection{LASSO}
\label{subsub:lasso}
If $\mathbf{\beta}$ is sufficiently sparse, a relaxation method called the ``LASSO (least absolute shrinkage and selection operator) method'' works well for variable selection in linear regression \cite{Tibshirani1996}. The performance of LASSO should be compared with that of the ES method. LASSO is formulated as the method of least squares with $L1$ regularization as
\begin{eqnarray}
\hat{\mathbf{\beta}}(\lambda)
= \mathrm{argmin}_{\mathbf{\beta}}
					\left\{
						\frac{1}{2}(\mathbf{y}- \mathbf{X}\mathbf{\beta})^\mathrm{T}\Sigma^{-1}
						(\mathbf{y}- \mathbf{X}\mathbf{\beta})
								+ \lambda ||\mathbf{\beta}||_1
						\right\}, 
\label{eq:LASSO}
\end{eqnarray}
where $\|\cdot\|_1$, called the ``$L1$-norm,'' is defined as $\|\mathbf{\beta}\|_1=\sum_i|\beta_i|$, and its coefficient $\lambda$ is called a ``regularization parameter.'' According to the notation of the indicator $\mathbf{c}$, it is convenient that the combination selected by LASSO with $\lambda$ is denoted by $\mathbf{c}(\lambda)$: $c_i(\lambda)=1$ if $\hat{\beta}_i(\lambda)\not=0$ and $c_i(\lambda)=0$ if not. We make some technical remarks. First, as preprocessing, the data $\mathbf{y}$ are standardized, and the explanatory variables $\mathbf{X}$ are centered such that $\beta_0$ is set to zero without loss of generality. The glmnet package in R was used for solving LASSO and its preprocessing \cite{Friedman2008}. Next, after LASSO, the non-zero coefficient values $\mathbf{\beta}$ are recalculated by the method of least squares to remove bias due to the $L1$-norm term \cite{Figueiredo2007}. Finally, regularization parameter $\lambda$ should be handled with much care. If $\lambda$ is set to a moderate value, LASSO suppresses some of the coefficients $\mathbf{\beta}$ to zero and leads to an appropriate sparse combination of explanatory variables. If not, however, an excessively sparse combination or a non-sparse combination can be obtained. We explain two ways of using LASSO with respect to regularization parameter $\lambda$.

\subsubsection{$\lambda$-optimization method}
In general, regularization parameter $\lambda$ is optimized by using the CVE. The CVE is calculated in the same way as explained in Subsubsection \ref{subsubsec:CV}. Given $\lambda$, $\hat{\mathbf{\beta}}(\lambda)$ is estimated from training data, and the value of CVE, denoted by $\mathrm{CVE}(\lambda)$, is calculated with validation data. 
Simply stated, the minimizer $\lambda_\mathrm{min}$ of CVE can be regarded as the optimal $\lambda$, but $\lambda_\mathrm{min}$ tends to select a variable combination that is not very sparse. Thus, a heuristic criterion called the ``one-standard-error (1SE) rule,'' by which the largest $\lambda_\mathrm{1SE}$ giving a larger CVE than the minimal CVE by at most the CVE's standard error is taken, is frequently used \cite{Murphy2012}. 

\subsubsection{$\lambda$-scan method}
The $\lambda$-scan method is inspired by the ES method. Instead of optimizing $\lambda$, the $\lambda$-scan method exhaustively searches whatever explanatory variable combination, $\mathbf{c}(\lambda)$, LASSO provides regardless of the value of $\lambda$. 
We calculate the FE and the CVE with respect to all the $\mathbf{c}(\lambda)$, and what minimizes each of them is taken as optimal. The $\mathrm{FE}(\mathbf{c}(\lambda))$ and $\mathrm{CVE}(\mathbf{c}(\lambda))$ can be calculated in the same way as in the ES method. LASSO plays the role of reducing the combination space, which is searched with the ES method. 

\section{Real data analysis}
\label{sec:real}
As mentioned in Section \ref{sec:intro}, using the ES-$K$ and AES-$K$ methods, we exhaustively searched sparse combinations of explanatory variables for type Ia supernova data in the Berkeley Supernova Database \cite{Silverman2012}. 
We used ES-$K$ method for values of $K=1,2,\dots,5$ and AES-$K$ method for values of $K=6,7$. 
In this paper, we denoted ES-$K$ method such as ES-$1$ method, in the case of $K=1$. 
We used $p=78$ samples of absolute magnitude, $\mathbf{y}$, and $N=276$ explanatory variables, $\mathbf{X}$. The explanatory variables consist of light-curve width, $x_1$, color, $c$, apparent magnitude, and the spectral data from $3500$ to $8500 \mathrm{\AA}$. We use three kinds of normalized spectra: continuum-normalized spectra ($134$ variables), this total-flux-normalized spectra ($134$ variables), and previously proposed flux ratios ($6$ variables) \cite{Silverman2012,Uemura2015}. We used $10$-fold CV in our analysis. 

\begin{figure*}[t]
  \begin{center}
   \includegraphics[width=1\linewidth]{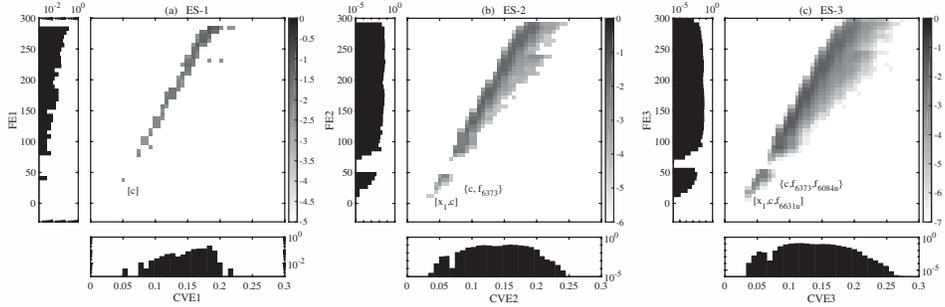}
  \end{center}
 \caption{Density of states for real data obtained using ES-$1$, -$2$, -$3$ methods. In each figure, horizontal and vertical histograms represent density of states corresponding to CVE and FE, respectively, and central figure shows two-dimensional density of states. These are expressed in logarithmic scales. Top combinations obtained with ES-$K$ method and $\lambda$-scan method are written in $[\cdot]$ and $\{ \cdot \}$, respectively. Two information criteria, namely, FE and CVE, lead to same top combination in each case of $K=1,2,3$.}
  \label{real_dos}
\end{figure*}

\begin{figure*}[t]
  \begin{center}
   \includegraphics[width=1\linewidth]{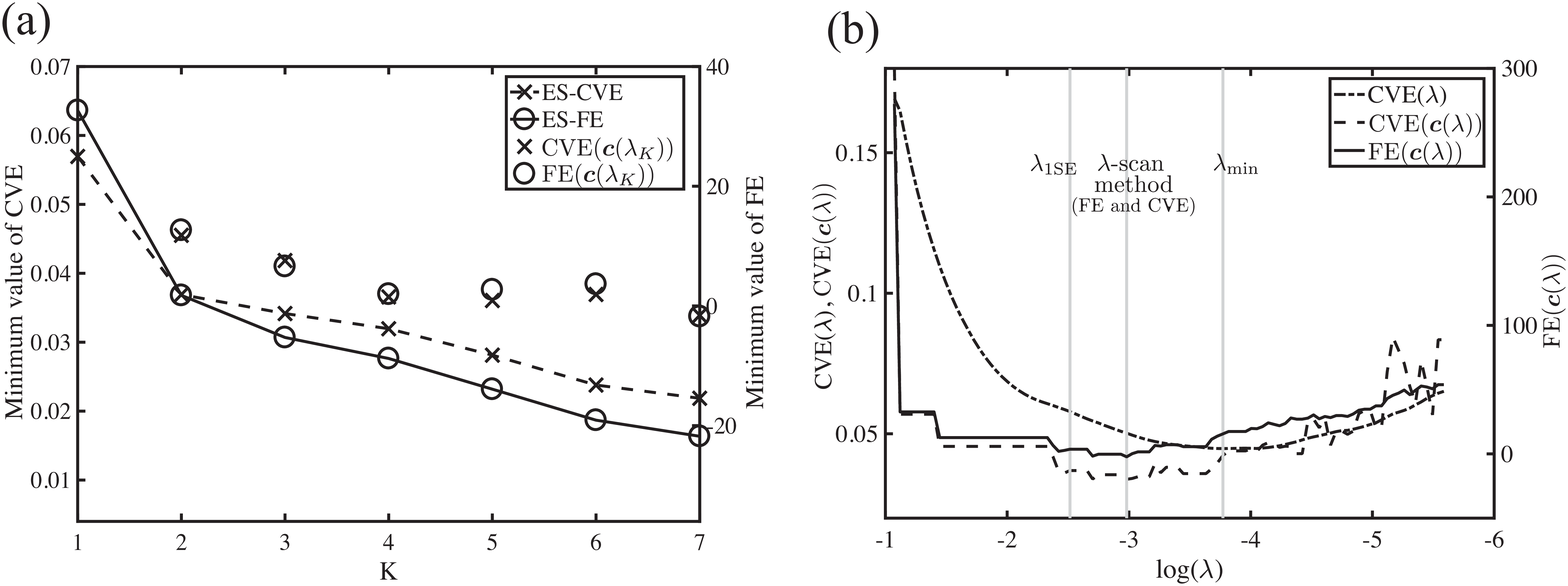}
  \end{center}
\caption{(a) Minimum value of FE and CVE for all combinations of $K$ explanatory variables for real data. We used ES-$K$ method for values of $K=1,2,\dots,5$ and AES-$K$ method for values of $K=6,7$. (b) Results of LASSO for real data. Figure shows dependency of CVE$(\lambda)$, CVE$({\mathbf c}(\lambda))$, and FE$({\mathbf c}(\lambda))$ on $\log(\lambda)$. Left two vertical lines represent $\lambda_\mathrm{1SE}$, $\lambda$ corresponding to results of $\lambda$-scan method based on FE, respectively. Center one shows $\lambda$ corresponding to results of $\lambda$-scan method based on CVE. Right one shows $\lambda_{\mathrm{min}}$.}
\label{real_min_lasso}
\end{figure*}

\subsection{Results of ES-$K$ and AES-$K$ methods}
\label{subsec:esk_aesk}
Figure \ref{real_dos} shows the density of states for real data obtained with the ES-$1$, -$2$, and -$3$ methods. Table \ref{ES_summary} indicates the top three variables for FE and CVE obtained with the methods. Almost all of the variables selected by the two criteria, FE and CVE, were the same, except for the third combination obtained with the ES-$3$ method. As shown in Fig. \ref{real_dos}(a) and Table \ref{ES_summary}, we found that the variable $\{c\}$ gave the minimal CVE and FE, which were significantly lower than those of other variables. As shown in Fig. \ref{real_dos}(b), for ES-$2$, there was a clustering structure with a low CVE and FE. The structure consisted of combinations of two variables including $\{c\}$ and another variable. The combination consisting of light-curve width $x_1$ and color $c$ had a remarkably lower CVE and FE than the other combinations of two explanatory variables, as shown in Fig. \ref{real_dos}(b) and Table \ref{ES_summary}. As shown in Fig. \ref{real_dos}(c), for the results for $3$ explanatory variables, we found that, similar to the results with ES-$2$, the combinations of three explanatory variables, including $\{x_1, c\}$, formed a cluster structure with a low FE and CVE. The FE and CVE for $\{x_1, c\}$ were higher than the minimal FE and CVE of the three explanatory variables, 
including $\{x_1, c\}$, formed a cluster structure with low FE and CVE. 
The FE and CVE for $\{x_1, c\}$ are higher than 
the minimal FE and CVE of the $3$ explanatory variables, including $\{x_1, c\}$. 
Since the ES-$2$ method agreed with the conventional understanding in astronomy \cite{Uemura2015}
that the absolute magnitude at maximum depends on the color and light-curve width, 
the ES-$K$ method is considered to be effective if the sparseness $K$ is known beforehand.

To estimate the sparseness $K$ from the data, we calculated the minimum values of FE and CVE for all combinations of $K=1,2,\dots,7$ explanatory variables for real data, as shown in Fig. \ref{real_min_lasso}(a). We used the ES-$K$ method for the values of $K=1,2,\dots,5$ and the AES-$K$ method for the values of $K=6,7$. For the REMC method, we set inverse temperatures $1/T_{\omega}$ ($\omega=1,2,\dots,15$) from $10^{0}$ to $10^{4}$ for CVE and from $10^{-3}$ to $10^{1}$ for FE, which are equally spaced on the logarithmic scale. The number of iteration steps for the REMC method was set to 100,000. The first half was taken as a burn-in period, and the second half was used to estimate the density of states with the multiple histogram method. We confirmed that the minimum values of FE and CVE derived with the ES-$K$ method corresponded to those with the AES-$K$ method for $K=1,2,\dots,5$. 
We found that the FE obtained with ES-$K$ and AES-$K$ methods reached a minimum at $K=6$ and  CVE monotonically decreased as $K$ increased, as shown in Fig. \ref{real_min_lasso}(a). 
This means that if the number of non-zero elements, $K$, is to be estimated from data, the ES-$K$ method does not support the conventional understanding in astronomy that $K=2$. 
In comparison with the FE for the uniform prior of $\mathbf{\beta}$, 
we also found that 
the FE for the estimated prior of $\mathbf{\beta}$ is more apt to be increased by an increase in the number of explanatory variable, $K$. 
Then, using the FE for the estimated prior of $\mathbf{\beta}$ 
for sparse variable selection, we can select a sparse variable combination. 

\begin{table*}[t]
\begin{center}
  \caption{Top three combinations for real data in terms of FE and CVE obtained using ES-$1,2,3$ methods and their CVE and FE}
   \begin{tabular}{p{2cm}p{1.5cm}p{3.5cm}p{1.5cm}p{1.5cm}}
   \hline
 Method & Ranking & Non-zero elements & CVE & FE 
  \\ \hline \hline \hline
  \item ES-$1$&
  \item 1st&
  \item  $c$&
  \item $0.057$&
  \item $40.7$
  \\ 
  \item &
  \item 2nd&
  \item $f_{6373}$&
  \item $0.078$& 
  \item $82.9$
  \\
  \item &
  \item 3rd&
  \item  $f_{6331}$ &
  \item  $0.080$  &
  \item  $83.5$
  \\ 
  \hline
  \item ES-$2$&
  \item 1st&
  \item  $x_1$, $c$&
  \item $0.037$&
  \item $13.2$
  \\ 
  \item &
  \item 2nd&
  \item $c$, $f_{6289n}$&
  \item $0.043$& 
  \item $23.8$
  \\
  \item &
  \item 3rd&
  \item  $c$, $f_{6373}$ &
  \item  $0.046$&
  \item  $24.1$
  \\ \hline
\item ES-$3$ &
  \item 1st&
  \item  $x_1$, $c$, $f_{6631n}$&
  \item  $0.034$&
  \item  $9.4$
  \\ 
  \item &
  \item 2nd&
  \item $x_1$, $c$, $f_{3752}$&
  \item $0.035$ &
  \item $13.1$ 
  \\
  \item &
  \item  3rd (FE) &
  \item  $x_1$, $c$, $f_{3703}$&
  \item  $0.037$ &
  \item  $13.4$
  \\
  \item &
  \item  3rd (CVE) &
  \item  $c$, $f_{6084n}$, $f_{6289n}$&
  \item  $0.036$ &
  \item  $14.0$
  \\ \hline
  \end{tabular}
  \label{ES_summary}
  \end{center}
\end{table*}

\begin{table*}[t]
\begin{center}
  \caption{ 
Non-zero elements obtained with the LASSO method 
 and their CVE($\mathbf{c}({\lambda})$) and FE$({\mathbf c}(\lambda))$ are listed. 
 Results of $\lambda$-optimization method (1SE) and $\lambda$-scan method (CVE and FE) are shown. 
}
   \begin{tabular}{p{2.5cm}p{5.5cm}p{1.5cm}p{1.5cm}}\hline
Method & Non-zero elements & CVE($\mathbf{c}({\lambda})$) &FE$({\mathbf c}(\lambda))$
  \\ \hline \hline \hline
  \item $\lambda$-optimization (1SE)&
  \item  $x_1$, $c$, $f_{6373}$,$f_{6084n}$, $f_{6289n}$, $f_{3780/4580}$&
  \item $0.037$& 
  \item $27.3$
  \\ 
  \item $\lambda$-scan (CVE)&
  \item $x_1$, $c$, $f_{3752}$, $f_{6373}$,$f_{6084n}$, $f_{6289n}$, $f_{6631n}$&
  \item $0.034$&
  \item $24.3$
  \\
  \item $\lambda$-scan (FE)&
  \item  $x_1$, $c$, $f_{6373}$,$f_{6084n}$&
  \item $0.037$&
  \item $19.8$
  \\ \hline
  \end{tabular}
  \label{LASSO_summary}
  \end{center}
\end{table*}

\subsection{Evaluation of results obtained with LASSO}
\label{subsec:eval_lasso}
In this section, we evaluate the performance of the LASSO methods when they were applied to the analysis for real data. Figure \ref{real_min_lasso}(b) shows CVE $(\lambda)$ used for the $\lambda$-optimization method and FE$(\mathbf{c}(\lambda))$ and CVE$(\mathbf{c}(\lambda))$ used for the $\lambda$-scan method against $\log\lambda$. Table \ref{LASSO_summary} shows non-zero explanatory variables selected by the LASSO methods. As previously reported \cite{Uemura2015}, we first use the $\lambda$-optimization method. According to the 1SE rule, the six explanatory variables listed in the top row of Table \ref{LASSO_summary} were selected as a sparse combination. The $\lambda$-optimization method failed to reproduce the conventional in astronomy understanding that the absolute magnitude at maximum depends on the color and light-curve width. 

We next used the $\lambda$-scan method to analyze real data. 
As shown in Table \ref{LASSO_summary}, the $\lambda$-scan methods with respect to the FE and the CVE provided the sparse combinations composed of four and seven explanatory variables, respectively. 
We made sure that the values of $\lambda$, which minimizes FE$(\mathbf{c}(\lambda))$ and CVE$(\mathbf{c}(\lambda))$, 
was comparable to $\lambda_\mathrm{1SE}$, as shown in Fig. \ref{real_min_lasso}(b). 
To compare the $\lambda$-scan method with the ES-$K$ method in detail, we classified all $\mathbf{c}(\lambda)$ into groups whose members were composed of $K$ non-zero explanatory variables and plotted the minimal FE, denoted by FE$(\mathbf{c}(\lambda_K))$, and the minimal CVE, denoted by CVE$(\mathbf{c}(\lambda_K))$, within each group in the cases of $K=1,2,\dots,7$ in Fig. \ref{real_min_lasso}(a). For $K\geq2$, the ES-$K$ method outperformed the $\lambda$-scan method in terms of both the FE and the CVE, as shown in Fig. \ref{real_min_lasso}(a). These results indicate that LASSO failed to find the $K$-sparse combination of explanatory variables that gives a minimum FE or CVE obtained with the ES-$K$ or AES-$K$ method. We stress that, especially in the case of $K=2$, the ES-2 method succeeded in extracting $\{x_1,c\}$, which is commonly believed to be important in astronomy \cite{Uemura2015}, whereas LASSO failed. We mapped the solution of the $\lambda$-scan method onto the density of states in Figs. \ref{real_dos}(a)-(c). In the case of $K=1$, the same result $\{c\}$ was obtained with the ES-$K$ and $\lambda$-scan methods. In the case of $K=2$, the solution of the ES-$K$ method, which follows the conventional understanding, composed the top cluster, and the solution of the $\lambda$-scan method was in the second best cluster. In the case of $K=3$, the solution of the $\lambda$-scan method provided a different solution from that of the ES-$K$ method, although both of them were in the top cluster. Thus, the density of states was powerful enough to give an overview of the relationship between the solutions of various methods. Consequently, the ES-$K$ method was more effective for sparse variable selection than LASSO. 

\section{Virtual measurement and analysis}
We compared our results stated in Section \ref{sec:real} with those of previous studies \cite{Uemura2015,Obuchi2016b,Kabashima2016}, which used LASSO with CVE and confirmed the conventional understanding that the absolute magnitude at maximum depends on the color and light-curve width, i.e., $\{x_1, c\}$. In contrast, our analysis using the ES and AES methods revealed that some combinations of explanatory variables have a higher performance both in terms of CVE and FE than $\{x_1,c\}$. Our results appear to be inconsistent with the previous work. 

To interpret the inconsistency, we conducted virtual measurement and analysis (VMA). What we call ``VMA'' is a numerical simulation using synthetic data in order to check whether a method works well with the available size of data. The fundamental problem for real data analysis is that, needless to say, no one knows what should be the result, namely, the truth. VMA addresses this problem by analyzing synthetic data, behind which the ``truth'' can be set, in exactly the same way as in a real situation. Using VMA, therefore, we examined what will happen in the process of our data analysis if we do not have enough data. We claim that VMA is completely different from numerical experiments using oversimplified artificial data or irrelevant databases and that it is significant in VMA to extract the essence of real measurement to build a virtual measurement model. A guiding principle of modeling has been discussed and presented previously as the three levels of data-driven science \cite{Igarashi2016}. 

Here, we summarize the procedure of our VMA. First, we built our virtual measurement model in accordance with previous astronomical work \cite{Uemura2015} and generated virtual data from the model. Then, we applied our proposed methods, namely, ES-$K$ and AES-$K$, and the LASSO methods to the virtual data. For the REMC method of AES-$K$, we set inverse temperatures $1/T_{\omega}$ ($\omega=1,2,\dots,15$) from $10^{0}$ to $10^{4}$ for CVE and from $10^{-3}$ to $10^{1}$ for FE, which are equally spaced on the logarithmic scale. Finally, we compared the virtual results with the real results stated in Section \ref{sec:real}. 

Our virtual model has $N=200$ explanatory variables $\mathbf{x}_\mu$ to imitate a real number, namely, $N=276$. Each element of $\mathbf{x}_\mu$ ($\mu=1,\dots,p$) is generated from $\mathcal{N}(0,1)$. Note that $\mathcal{N}(\mu,\sigma^2)$ represents a normal distribution with mean $\mu$ and variance $\sigma^2$. According to the conventional understanding \cite{Uemura2015}, the true number of non-zero explanatory variables is set to two. More specifically, $\mathbf{\beta}$ has 200 elements, and $\beta_1$ and $\beta_2$ are two non-zeros. The non-zero coefficients $\beta_1$ and $\beta_2$ are generated from $\mathcal{N}(0,\sigma_\beta^2)$. A virtual dataset $\mathbf{y}$ is obtained by virtual measurement: 
\begin{eqnarray}
	\mathbf{y}=\mathbf{X}\mathbf{\beta}+\mathbf{\epsilon}.
\end{eqnarray}
Each noise component $\epsilon_\mu$ is assumed to follow $\mathcal{N}(0,\sigma_\epsilon^2)$. To make the virtual signal-to-noise ratio comparable to the real value \cite{Uemura2015}, we set $\sigma_\beta^2=1$ and $\sigma_\epsilon^2=0.1$. We conducted VMA for $p=700,50,30$ to investigate the effect of sample size.

\begin{figure*}[t]
\begin{center}
\includegraphics[width=1\linewidth]{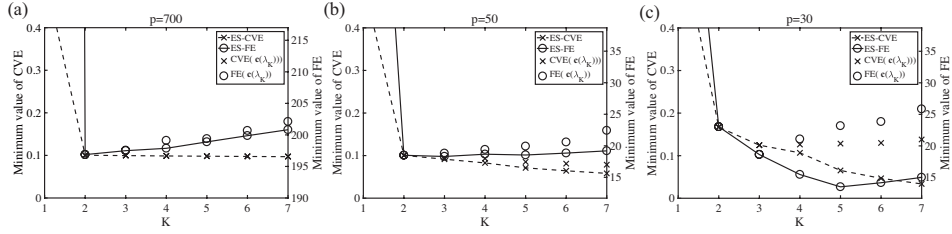}
\caption{Results of VMA. Only two non-zero elements were set in $N$-dimensional coefficient vector, and minimal FE or CVE was derived using LASSO and ES-$K$ (or AES-$K$) methods with $p=700,50,30$ and $N=200$. $K=1,2,\dots, 7$ explanatory variables were selected using $\lambda$-scan method, and FE$(\mathbf{c}(\lambda_K))$ and CVE$(\mathbf{c}(\lambda_K))$ were calculated, as shown by black circles and cross marks, respectively. Minimal FE and CVE with $K=1,2,\dots,7$ explanatory variables and ES-$K$ or AES-$K$ methods are represented by set of black circles connected by solid lines or set of cross marks connected by dotted lines, respectively. ES-$K$ method was used for values of $K=1,2,3$, and AES-$K$ method was used for values of $K=4,\dots,7$.}
\label{vma_ES_LASSO}
\end{center}
\end{figure*}

Figure \ref{vma_ES_LASSO} shows the VMA results corresponding to those of real data analysis in Fig. \ref{real_min_lasso}(a). If the dataset was sufficiently large ($p=700$), the FE obtained with the ES-$K$ method reached a minimum at $K=2$, which is correct, as shown in Fig. \ref{vma_ES_LASSO}(a). 
However, CVE obtained with the ES-$K$ method monotonically decreased as $K$ increased. This means that FE is more suitable for selecting sparse variables than CVE. As the number of samples $p$ was reduced ($p=50,\; 30$), the FE obtained with the ES-$K$ method reached a minimum at $K>2$, which is not correct, as shown by the set of black circles connected by solid lines in Figs. \ref{vma_ES_LASSO}(b) and (c). These results imply that variable selection fails when $p$ is below a threshold value and that there is a phase transition in variable selection with respect to $p$. 

We also discuss the results of VMA using the LASSO methods. Using the $\lambda$-scan method, we selected a $K$-sparse combination of explanatory variables for $K=1,2,\dots,7$. We then plotted FE$(\mathbf{c}(\lambda_K))$ and CVE$(\mathbf{c}(\lambda_K))$, shown as black circles and cross marks, respectively, in Fig. \ref{vma_ES_LASSO}. Figure \ref{vma_ES_LASSO}(a) shows that when the dataset was sufficiently large ($p=700$), the results obtained using the $\lambda$-scan method with FE reached a minimum at $K=2$. This means that LASSO lead to the true number of non-zero explanatory variables as well as the ES-$K$ method. As $p$ decreased ($p=50,30$), however, the FE and CVE minimized by using the $\lambda$-scan method did not coincide with those obtained using the ES-$K$ and AES-$K$ methods in the case of $K\geq3$. 

We compare Fig. \ref{real_min_lasso}(a) with Fig. \ref{vma_ES_LASSO}. In the case of a sufficiently large $p$, both the ES-$K$ method and $\lambda$-scan method based on the FE were successful in selecting an appropriate $K$ because there was a notch of FE at $K=2$, and the CVE based methods were not successful because the minimized CVE decreased as $K$ increased, 
as shown in Fig. \ref{vma_ES_LASSO}(a). 
However, when $p$ was reduced, the notch of FE at $K=2$ disappeared, and FE decreased more rapidly against $K$ than CVE as shown in Fig. \ref{vma_ES_LASSO}(c). This behavior of FE and CVE is similar to that in Fig. \ref{real_min_lasso}(a). Therefore, it is regarded that the situation of real data analysis corresponds to the case of data shortage, due to which the ES-$K$ method fails in selecting the combination composed of color and light-curve width $\{x_1,c\}$ as explanatory variables of the absolute magnitude at maximum. If the optimal sparseness is to be estimated from only data, much more data are needed. In addition, these VMA results indicate that the ES-$K$ and AES-$K$ methods are more credible when selecting explanatory variables by optimizing a certain information criterion than the LASSO methods, such as 
Least angle regression(LARS)-LASSO \cite{Efron2004} and the $\lambda$-scan method. 
\section{Conclusion}
We proposed ES-$K$ and AES-$K$ methods for selecting sparse variables in linear regression. In the ES-$K$/AES-$K$ framework, we assume that the optimal combination of explanatory variables is $K$-sparse and evaluate all $K$-sparse combinations at the expense of exponentially increasing computational complexity. We also applied the methods to type Ia supernova data \cite{Silverman2012,Uemura2015} and compared them with LASSO methods. According to previous studies, it has been understood that the absolute magnitude at maximum depends on the color and light-curve width \cite{Uemura2015,Obuchi2016b}. Our analysis has shown that, given a number of explanatory variables, the ES-$K$ method succeeds in sparse variable selection, whereas LASSO does not. In addition, the ES-$K$/AES-$K$ framework provides the density of states of explanatory variable combinations. By mapping the solutions of LASSO and other various approximate methods onto the density of states, their performance can be systematically evaluated. Therefore, we claim that the ES-$K$/AES-$K$ framework is important for sparse variable selection.

We also revealed that the ES-$K$ method leads to a combination of explanatory variables different from those of the conventional understanding in astronomy. We discussed this inconsistency using VMA. The advantage of VMA is that it enables us to control the truth behind data once a virtual measurement model is good enough. By changing the size of data in VMA, we examined whether the size reflects the reality well. In the case of a large data size, we showed that the ES-$K$ method based on the FE successfully confirmed the conventional understanding. However, as the data size was reduced, more explanatory variables than needed were selected. Then, we argued that, in the analysis of type Ia supernova data, the size of the current dataset is not sufficient enough to derive a faithful conclusion. The phase transition of sparse variable selection with respect to the amount of data is of great interest. It is necessary to understand the mathematical mechanism of this kind of phenomena, which will promote research on judging whether given data are enough or not. VMA conducted in combination with the ES-$K$/AES-$K$ framework will be a key technology for dealing with such problems. 

\section*{acknowledgment}

We thank Dr. J. M. Silverman and Dr. A. V. Filippenko of the University of California 
for providing the Berkeley Supernova Database. 
This work was supported 
by JSPS Grants-in-Aid for Scientific Research on Innovative Areas 
(Grant P25120001, JP25120008, JP25120009) 
and 
Grants-in-Aid for JSPS Fellows (No. 15J07765).

\bibliographystyle{tieice}
\bibliography{myrefs}

\end{document}